\documentclass[sigconf,authorversion,nonacm]{acmart}
\AtBeginDocument{%
  \providecommand\BibTeX{{%
    \normalfont B\kern-0.5em{\scshape i\kern-0.25em b}\kern-0.8em\TeX}}}

\setcopyright{acmcopyright}
\copyrightyear{2023}
\acmYear{2023}
\acmDOI{XXXXXXX.XXXXXXX}

\acmConference[WWW ’24 Companion]{ACM Web Conference 2024}{MAY 13 - 17, 2024}{Singapore}
\acmPrice{15.00}
\acmISBN{978-1-4503-XXXX-X/18/06}

\usepackage{multirow}

\begin{document}

\title{Minimizing Factual Inconsistency and Hallucination in Large Language Models}

\author{Muneeswaran I}
\email{muneeswaran.i@quantiphi.com}
\affiliation{
  \institution{Applied Research, Quantiphi}
  \country{India}
}

\author{Shreya Saxena}
\authornote{Equal contribution}
\email{shreya.saxena@quantiphi.com}
\affiliation{
  \institution{Applied Research, Quantiphi}
  \country{India}
}

\author{Siva Prasad}
\authornotemark[1]
\email{siva.prasad@quantiphi.com}
\affiliation{
  \institution{Applied Research, Quantiphi}
  \country{India}
}

\author{M V Sai Prakash}
\authornotemark[1]
\email{mukkamala.prakash@quantiphi.com}
\affiliation{
  \institution{Applied Research, Quantiphi}
  \country{India}
}

\author{Advaith Shankar}
\email{advaith.shankar@quantiphi.com}
\affiliation{
  \institution{Applied Research, Quantiphi}
  \country{India}
}

\author{Varun V}
\email{varun.v@quantiphi.com}
\affiliation{
  \institution{Applied Research, Quantiphi}
  \country{India}
}

\author{Vishal Vaddina}
\authornote{Corresponding authors}
\email{vishal.vaddina@quantiphi.com}
\affiliation{
  \institution{Applied Research, Quantiphi}
  \country{Canada}
}
\author{Saisubramaniam Gopalakrishnan}
\authornotemark[2]
\email{gopalakrishnan.saisubramaniam@quantiphi.com}
\affiliation{
  \institution{Applied Research, Quantiphi}
  \country{India}
}

\renewcommand{\shortauthors}{Muneeswaran, et al.}

\begin{abstract}
Large Language Models (LLMs) are widely used in critical fields such as healthcare, education, and finance due to their remarkable proficiency in various language-related tasks. However, LLMs are prone to generating factually incorrect responses or "hallucinations," which can lead to a loss of credibility and trust among users. To address this issue, we propose a multi-stage framework that generates the rationale first, verifies and refines incorrect ones, and uses them as supporting references towards generating the answer. 
The generated rationale enhances the transparency of the answer and our framework provides insights into how the model arrived at this answer, by using this rationale and the references to the context.
In this paper, we demonstrate its effectiveness in improving the quality of responses to drug-related inquiries in the life sciences industry. Our framework improves traditional Retrieval Augmented Generation (RAG) by enabling OpenAI GPT-3.5-turbo to be 14-25\% more faithful and 16-22\% more accurate on two datasets. Furthermore, fine-tuning samples based on our framework improves the accuracy of smaller open-access LLMs by 33-42\%, and competes with RAG on commercial models. \footnote{Preprint version}
\end{abstract}

\keywords{Large Language Models, Hallucination,
Information Retrieval}



\maketitle

\section{Introduction}
Large Language Models (LLMs) have revolutionized natural language processing, enabling impressive advancements in various tasks such as question-answering \cite{liu2023summary,lee2020generating}, summarization \cite{loem2022extraphrase, liu2023summary}, language translation \cite{liu2023summary}, sentiment analysis \cite{yang2020finbert}, and  classification \cite{liu2023summary, mayer2023prompt}. Consequently, LLMs have become a popular area of research in recent years, and their potential applications in various domains have garnered significant attention from both academia and industry. However, as these models have grown in scale and complexity, they have also exhibited an alarming propensity for generating inconsistent or fabricated information, a phenomenon commonly referred to as \textit{hallucination}\cite{zhang2023hallucination}. This issue raises critical concerns, particularly in applications where accuracy and reliability are paramount. 
Hence, developing robust techniques to mitigate the issue of factual inconsistencies and hallucinations in LLMs is of prime importance.

Retrieval Augmented Generation (RAG) \cite{lewis2020retrieval} has shown impressive results in generating grounded responses for natural language queries, however, industrial applications often require a higher degree of transparency, where the generated response must be traced back to a source, most often available in internal databases. From an industry perspective, we consider certain key points that are applicable: (i) Enterprise user interactions are mostly in
the form of question and answers, (ii) Primary business use cases extend the aspect of traditional information retrieval with a need for an informed response for their task and domain, (iii) The required responses are obtainable from factually curated knowledge repositories, constituting internal documents (unstructured) and databases (structured), or their derivatives (knowledge graphs), with an occasional need for web retrieval, (iv) Reducing hallucination related to misrepresentation of facts in the response is of prime importance, and, (v) Ability to trace back to the source of the response as a means of
interpretability and user trust. Considering the above points, relying on the vanilla version of RAG alone will not be sufficient to cater to the requirements of these applications.

In the life sciences industry, Pharmacovigilance \cite{meyboom1999pharmacovigilance} plays a crucial role in public health by ensuring the post-market safety of pharmaceuticals. This discipline involves collecting, analyzing, and reporting data on adverse events, allowing for informed regulatory
decisions. Pubmed 
articles serve as a critical source of information for drug-related inquiries. However, the vast volume of information available can be challenging to navigate, making it difficult to identify relevant information accurately. Using LLMs can be a practical solution here to mitigate some of the challenges and ensure efficient data management. This can help enhance the efficiency and accuracy of pharmacovigilance activities, leading to improved drug safety and better public health outcomes.

Our multi-stage framework takes all the aforementioned key points into consideration. 
In our work, we:
\begin{itemize}
\item Propose a multi-stage framework that explicitly generates rationales, verifies and refines incorrect ones, and uses them as supporting references to generate accurate responses. The framework enhances transparency and provides users with insights into how the model arrived at the final responses, including references to the context,
\item Demonstrate the effectiveness of our framework in the biomedical industry by evaluating it on two datasets: (i) PubmedQA\cite{Jin2019PubMedQAAD}, a public question-answering dataset, and (ii) Adverse-Effect-QA (AEQA), an internal expert-curated dataset focusing on adverse drug reactions and medical conditions specific to pharmacovigilance. Our framework outperforms RAG by {14.1\%, 15.82\%} in faithfulness and accuracy on PubmedQA, and by {25.04\%, 21.66\%} on our AEQA dataset,
\item Show that a fine-tuned Llama2-7B with samples following our framework is at least 30\%-40\% better than non fine-tuned versions and competes with RAG on larger commercial options.
\end{itemize}

While our work in this paper focuses on improving the quality of responses to drug-related inquiries for a life science-based application, our framework's effectiveness extends to other industry applications, making it a versatile and impactful tool for organizations seeking to enhance the accuracy and transparency behind the rationale of the generated LLM responses.

\section{Background}
\subsection{Related Work on LLM Hallucination}
Large Language Models (LLMs) have emerged as a transformative technology in Natural Language Processing (NLP) and Natural Language Generation (NLG), surpassing traditional methods in both scale and performance. Despite increased model and data scale, state-of-the-art LLMs still struggle with factual errors \cite{ouyang2022training, ji2023survey}.
This has been largely attributed to LLMs memorizing vast amounts of knowledge in their parameters through pre-training and fine-tuning, and this phenomenon is referred to as parametric knowledge \cite{roberts2020much, mallen2023not}. This stored knowledge can lead to hallucinations if the underlying information either gets outdated, or the LLM hidden activations lead it towards incorrect responses during inference \cite{xie2023adaptive}. \cite{zhang2023siren} categorizes hallucination into input, context, and fact-conflicting. In particular, fact-conflicting hallucination occurs when LLMs generate text or information that contradicts established knowledge. To mitigate this, various methods can be applied during pre-training, fine-tuning (SFT or RLHF), or inferencing (decoding strategies, external knowledge, exploiting uncertainty, etc.). In the latter, existing approaches either supplement external knowledge \cite{mialon2023augmented} during generation time \cite{borgeaud2022improving} or use it for post-hoc correction \cite{cao2020factual}. Existing approaches such as Retrieval-Augmented Generation (RAG) \cite{lewis2020retrieval} augment LLM input with retrieved relevant passages and demonstrations to utilize the inherent capability of In-Context Learning (ICL) \cite{dong2022survey}. While RAG and its In-Context variant \cite{ram2023context} have reduced factual errors in knowledge-intensive tasks, industrial applications also require explanations or the rationale behind the responses
 coupled with transparency, where the generated response must be traceable to the source, often available in their internal data stores.

\subsection{Pharmacovigilance as a case study}
\begin{figure*}
\includegraphics[width=0.95\textwidth]{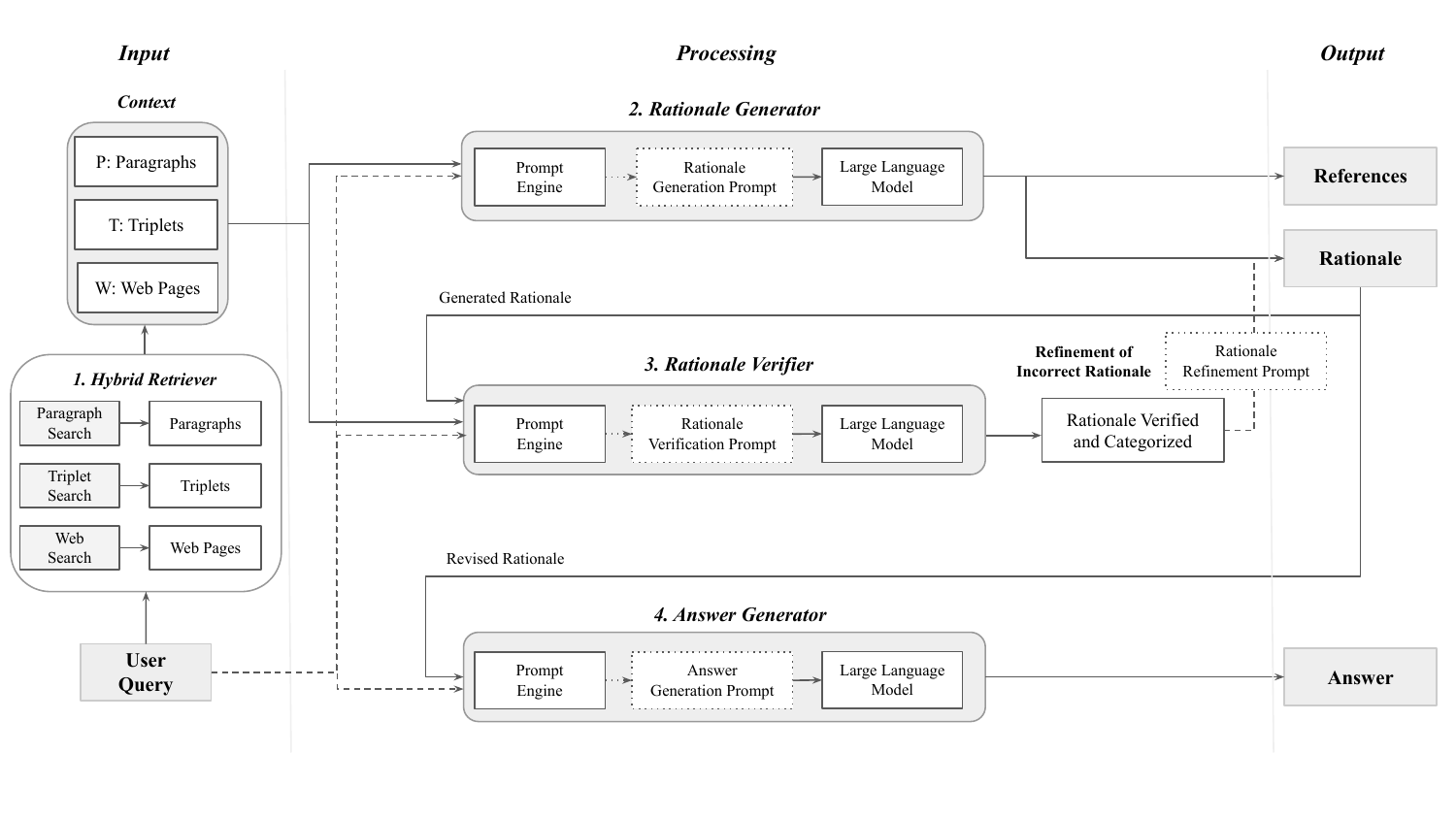}
  \vspace{-1.cm}
  \caption{Our proposed multi-stage approach showcasing various components - Hybrid Retriever, Rationale Generator, Rationale Verifier and Refiner}
\end{figure*}

Pharmacovigilance encompasses the amalgamation of scientific inquiry and associated activities focused on the identification, comprehension, and mitigation of potential drug-related adverse effects or problems. Its objective is to ensure the judicious and secure administration of medications, thereby upholding both public health and the well-being of patients. A pivotal facet of pharmacovigilance lies in the aggregation and scrutiny of safety-related information pertaining to pharmaceuticals \cite{meyboom1999pharmacovigilance}. The underlying principle is straightforward: in order to enhance patient outcomes, the availability of high-caliber Adverse Drug Reaction (ADR) reports is of paramount importance. The generation of such reports hinges upon proficient data collection methodologies and the capability to analyze them. The burgeoning volume of accessible data points represents an additional challenge for conventional Pharmacovigilance. Numerous authors have articulated their apprehensions in a publication presented by experts affiliated with the International Society of Pharmacovigilance (ISoP) \cite{2017}. The expanding global population is anticipated to lead to an exponential surge in the inflow of ADR-related data. Unfortunately, the methodologies employed in traditional Pharmacovigilance are ill-equipped to manage such a prodigious volume of data.
In essence, conventional Pharmacovigilance is projected to entail higher costs without effectively handling the forthcoming surge in data. As many seek alternative approaches, it appears that Large Language Models (LLM) offer potential solutions to the predicaments confronting traditional Pharmacovigilance. The integration of LLMs in Pharmacovigilance is instrumental for efficient data processing. LLMs excel in analyzing large volumes of data, particularly free-form text data prevalent in pharmaceuticals and healthcare, ensuring more accurate data processing. Moreover, their incorporation introduces a higher level of automation in routine tasks, allowing professionals to focus on higher-value objectives leading to a more refined risk-benefit assessment for both new and existing drugs in the market. Even though they exhibit notable advantages, Large Language Models (LLMs) still have the limitation of occasionally producing factually inconsistent responses. In high-stakes domains such as the pharmaceutical industry, it remains crucial to address this concern.


Our work is designed to address the issue of fact-conflicting hallucinations during response generation. As opposed to other sophisticated works that train or fine-tune their own LLMs, our framework benefits existing LLMs without explicit tweaks. While our approach is generalizable, we selected pharmacovigilance in life sciences as our application domain, given its critical role in ensuring the safety of pharmaceutical products and the importance of generating accurate responses with appropriate references in this field.

\section{Methodology}
Our proposed multi-stage framework (termed Factual Evidence or FE for short) comprises five components: (i) Hybrid Retriever, which retrieves relevant information (context) based on the user query from various data sources such as documents, knowledge graphs, and the internet, (ii) Prompt Engine, which takes the context and user query as input and creates concise prompts with instructions for each component, (iii) Rationale Generator, which uses the context and the user query to generate rationale (explanation) with supporting evidence, (iv) Rationale Verifier, which verifies the factual accuracy of the rationale and whether it is explicitly or implicitly mentioned in the context, and Rationale Refiner, which revises or refines incorrect rationale based on the context, and, (v) Answer Generator, which utilizes the verified and refined rationale to facilitate the LLM to generate the final descriptive response ensuring factual accuracy of the answers. We describe each component in Figure 1 in detail.

\vspace{-0.05cm}
\subsection{Hybrid Retriever}
The Hybrid Retriever component handles both text and graph (triplets) to efficiently retrieve relevant information from diverse sources and formats, including indexed paragraph chunks from documents, knowledge graphs, and the web.  The retrieval system consists of multiple retrievers, each specialized in different search methods and data types, including semantic vectors, lexical, and knowledge graphs. Each retriever can be used separately or in combination to retrieve granular-level information from indexed data stores. The retrieval system also includes a re-ranker that retains only the top information based on the relevancy score. This forms the \textit{Context} that is passed as input to the LLM along with the user query and the prompt.

\subsubsection{Paragraph Search}
\label{paragraph-retrieval}
Paragraph Search retrieves information in the form of paragraph chunks $P = \{p_1, p_2, ..., p_m\}$ that are split and indexed from a set of internal documents $D = \{d_1, d_2, ..., d_n\}$.  Given a user query $q$, we employ both semantic and lexical techniques, to retrieve the top $k$ relevant paragraphs.

\textbf{Semantic Search:} We employ an encoder-only language model as the retriever to obtain embeddings of the query $q$ and perform a semantic vector search over the data store containing embedding vectors of all the paragraph chunks. Cosine similarity is used to retrieve the top-$k$ relevant paragraphs that are most semantically similar to the user's query, i.e., $P_{sem} = {p_{sem_1}, p_{sem_2}, ..., p_{sem_k}}$, where $p_{sem_i}$ is the $i^{th}$ most semantically similar paragraph chunk to $q$.

\textbf{Lexical Search:} Here, we retrieve the top-$k$ relevant paragraph chunks that contain the exact word or phrase in the user's query, i.e., $P_{lex} = {p_{lex_1}, p_{lex_2}, ..., p_{lex_k}}$, where $p_{lex_i}$ is the $i^{th}$ paragraph chunk that has a match to $q$.

Once both search methods retrieve their top-$k$ relevant paragraph chunks, we use a re-ranker to provide a relevancy score for each retrieved chunk with respect to $q$. The re-ranker retains only the top-$P_k$ paragraph chunks based on the decreasing relevancy scores. The weightage of lexical to semantic contribution to $P_k$ is decided based on the nature of the application.
By utilizing both Semantic Search and Lexical Search in parallel, we retrieve the most relevant paragraph chunks that contain both semantically similar and exact matches to the user's query.

\subsubsection{Triplet Search}
Triplet Search retrieves relevant information in the form of triplets from either Knowledge Graphs (KG) if constructed beforehand, or from paragraphs that are transformed into triplets during retrieval. In the former, i.e. given a user query $q$ and a Knowledge Graph $KG$, we retrieve the top $k$ triplets $T = {t_1, t_2, ..., t_k}$ that are most relevant to the query $q$ by utilizing a sub-graph retrieval algorithm \cite{zhang2022subgraph} and traversing through $KG$, taking into account the semantic relationships between the entities in $KG$ and the query $q$.
In the latter case, i.e. given a user query $q$ and a set of paragraph chunks $P$, we retrieve the top $k$ triplets $T = {t_1, t_2, ..., t_k}$ by processing the retrieved paragraph chunks on the fly into entities and relations. We replace all references of a sentence with their respective entity mentions, extract triplets, link entity mentions to their original entities, and canonicalize the resulting triplets. This reduces the token count for the context while maintaining a similar performance to using the original paragraph chunks.

\subsubsection{Web Search}
Web Search retrieves relevant information in the form of natural language text from the internet using selective search engine and social media APIs. Given a user query $q$, we retrieve the text of the top $k$ web pages and API responses $W = {w_1, w_2, ..., w_k}$ that are most relevant to the query $q$. We also offer the choice to users to provide a list of web pages that are whitelisted and should be exclusively used during the retrieval process. This helps to narrow down the search request. The obtained text is processed into paragraph chunks on the fly using the same method as described in Section \ref{paragraph-retrieval}. We ensure to give proper citations to the original webpage and API source when we provide the response. Additionally, we do not retain any information from Web Search in our data stores.

Overall, the Hybrid Retriever combines the results of the Paragraph, the Triplet, and the Web searches to produce a comprehensive set of relevant information, i.e. the \textit{Context} $C = \sum_{p_i \in P_k} w_i + \sum_{t_i \in T_k} t_i + \sum_{w_i \in W_k} p_i$, $+$ denoting concatenation per line, forming the primary source of knowledge for grounding the LLM towards generating factual responses along with the rationale and references. Each information of the context is associated with at
least one identifier (PID-, TID-, WID-*) for (paragraph, triplet, and
web) respectively.

\subsection{Prompt Engine}
The Prompt Engine is a shared component responsible for converting the user query $q$, the retrieved context in the form of text and triplets $C$, along with any of the available other component outputs, such as the generated rationale $R$, and verification statements $V$, into a concise, predefined prompt instruction $\textbf{p}$ template. The prompts are then passed to the LLM at each stage to generate the corresponding output. 

\subsection{Rationale Generator}
The Rationale Generator generates intermediate rationale (explanations) for a given query using the retrieved context. It plays a crucial role in guiding the LLM toward a factually grounded answer. It also allows the user to trace back the answer to the key part of the context, making it evidence-based and trustworthy.

Given a query $q$ and context $C = {c_1, c_2, ..., c_k}$ as input, the Prompt Engine generates a prompt instruction $\textbf{p}$ which is then fed to the LLM to identify and generate the rationale $R = {r_1, r_2, ..., r_k}$ on both implicit and explicit reasoning over each retrieved part of the context.
This is done to ensure a one-to-one mapping between the generated set of rationale and each retrieved part of the context in $C$. Each explanation in $R$ is associated with at least one identifier (PID-, TID-, WID-*) depending on the retrieved source.
\begin{align*}
\text{Rationale}\ R = \text{PromptEngine}(q,C,\textbf{p}(rationale\_generator\_prompt))
\end{align*}

\subsection{Rationale Verifier and Refiner}
\begin{table*}[t]
\caption{Comparison of RAG+FE with baselines on PubMedQA and AEQA Datasets} 
\centering
\begin{tabular}{@{}c c ccc ccc@{}}
\toprule
\multirow{3}{1pt}{Method} & \multirow{3}{1pt}{Model} & \multicolumn{3}{c}{PubMedQA} & \multicolumn{3}{c}{AEQA} \\ \cmidrule(l){3-8}

 &  & Faithful (\%) & Grade Score (out of 5)  & Accuracy (\%) & Faithful (\%) & Grade Score (out of 5) & Accuracy (\%) \\ \midrule
None & & 56.70 & 2.77 & 55.92 & 44.40 & 3.54 & 65.21 \\
RAG & GPT-3.5 Turbo & 72.40 & 3.13 & 72.67 & 58.20 & 3.28 & 75.19 \\
RAG+FE &  & \textbf{86.50} & \textbf{4.13} & \textbf{88.49} & \textbf{83.24} & \textbf{4.34} & \textbf{96.85} \\
\midrule

None & & 8.82 & 1.40 & 10.30 & 6.82 & 1.21 & 3.71 \\
RAG & LLama-2-7B & 7.19 & 1.38 & 8.92 & 3.78 & 1.23 & 5.64 \\
RAG+FE &  & \textbf{27.19} & \textbf{2.46} & \textbf{30.80} & \textbf{31.29} & \textbf{2.92} & \textbf{35.0} \\
\midrule
None & LLama-2-7B-FT & 14.97 & 1.73 & 21.10 & 24.18 & 2.53 &  20.70 \\
RAG+FE &  & \textbf{70.8} & \textbf{3.96} & \textbf{72.23} & \textbf{76.8} & \textbf{3.72} & \textbf{68.18}   \\ \bottomrule
\end{tabular}
\label{tab:main-table}
\end{table*}
\subsubsection{Rationale Verifier}
\label{verifier}
The Rationale Verifier evaluates the factualness and relevance of the generated rationale with respect to the query and the context. Each statement in the rationale is classified into one of the below based on its relevance to the respective identified part of the context: 
\begin{itemize}
    \item "CORRECT-EXPLICIT" 
    \item "CORRECT-IMPLICIT"
    \item "CORRECT-ADDITIONAL\_INFO"
    \item "INCORRECT-FALSE\_INFO" 
    \item "INCORRECT-DEVIATING\_INFO"
    \item "INCORRECT-ILLOGICAL"
\end{itemize}
Given a query $q$, context $C = {c_1, c_2, ..., c_k}$, and the rationale  $R = {r_1, r_2, ..., r_k}$ as input, the Prompt Engine generates a prompt instruction $\textbf{p}$ which is then fed to the LLM to verify and classify the rationale $R$ into verification statements $V = {v_1, v_2, ..., v_k}$.
\begin{align*}
\text{Verifier}\ V = \text{PromptEngine}(q,C,R,\textbf{p}(rationale\_verifier\_prompt))
\end{align*}
The Rationale Verifier classifies every statement in the rationale and offers a comprehensive justification for its categorization. Additionally, it determines a verdict using the assessment criteria, labeling the statement as either "CORRECT" or "INCORRECT." This aids in filtering out the incorrect statements and sending them to the Rationale Refiner.

\subsubsection{Rationale Refiner}
The Rationale Refiner reviews and improves erroneous parts of the rationale by utilizing the justification received from the Rationale Verifier as feedback. It assesses the feedback along with the query and context provided and integrates it into the revised rationale. The Rationale Refiner is an optional component and may be skipped if there are no INCORRECT labels from the Rationale Verifier.

Given a query $q$, context $C = {c_1, c_2, ..., c_k}$, rationale  $R = {r_1, r_2, ..., r_k}$, and verification statements $V = {v_1, v_2, ..., v_k}$ as input, the Prompt Engine generates a prompt instruction $\textbf{p}$ which is then fed to the LLM to revise the incorrect statements from $R$ using the feedback from $V$ to provide the revised rationale $R' = {r'_1, r'_2, ..., r'_k}$.
\begin{align*}
\text{Refined Rationale}\ R' = \text{PromptEngine}(q,C,R,V, \\
\textbf{p}(rationale\_refiner\_prompt))
\end{align*}

\subsection{Answer Generator}
The Answer Generator utilizes the meticulously verified and refined rationale to generate a factually grounded answer for the given query. Since we already have the rationale as support, we do not require the full context to generate the answer. The Answer Generator also incorporates the context identifiers as citations along with the rationale (explanation) behind each citation, allowing users to easily cross-reference them with parts of the answer and verify the sources used to generate the answer.
Given a query $q$ and the refined rationale  $R' = {r'_1, r'_2, ..., r'_k}$ as input, the Prompt Engine generates a prompt instruction $\textbf{p}$ which is then fed to the LLM to provide the final answer $A$ along with citations for cross-referencing parts of the answer to the context along with the rationale.
\begin{align*}
\text{Answer}\ A = \text{PromptEngine}(q,R', \textbf{p}(answer\_generator\_prompt))
\end{align*}

\section{Experiments}
We conducted a set of experiments to substantiate the effectiveness of our work. These experiments encompassed a methodical comparison with established baselines to assess the performance and efficacy of our approach.
 Our approach to retrieval is similar to that described in \cite{10020725}. However, our focus is on using the retrieved information to generate effective LLM responses with rationale and not on the hybrid retrieval component itself. We omit the experimental details and results related to the hybrid retrieval component. We compare our proposed approach - FE with RAG on both commercial (OpenAI GPT 3.5-turbo and open-access (LLama-2 \cite{touvron2023llama}) LLMs.


\subsection{Datasets}
\subsubsection{PubmedQA}
PubMedQA \cite{jin2019pubmedqa} is a  biomedical question answering (QA) dataset sourced from PubMed abstracts. Its primary objective is to furnish responses to research inquiries by utilizing the corresponding abstracts. This dataset encompasses 1,000 meticulously curated, expert-annotated QA instances. Each instance comprises of four essential components:  a question, a context denoting the corresponding abstract, an extensive answer, and a response categorized as either affirmative, negative, or uncertain.

\subsubsection{Adverse Effect Question Answering (AEQA) Dataset}
We curated an Adverse Effect Database containing 19,432 documents related to 300 drugs through the utilization of commercially available APIs in the biomedical domain. Subsequently, we created an evaluation dataset in collaboration with Subject Matter Experts, who meticulously selected intricate queries and their corresponding answers from the corpus. This finalized AEQA dataset encompasses 200 Question-Answer pairs pertaining to 40 distinct drugs, each accompanied by a context with an average token length of 1846. 

\begin{figure*}[t]
\includegraphics[width=\textwidth]{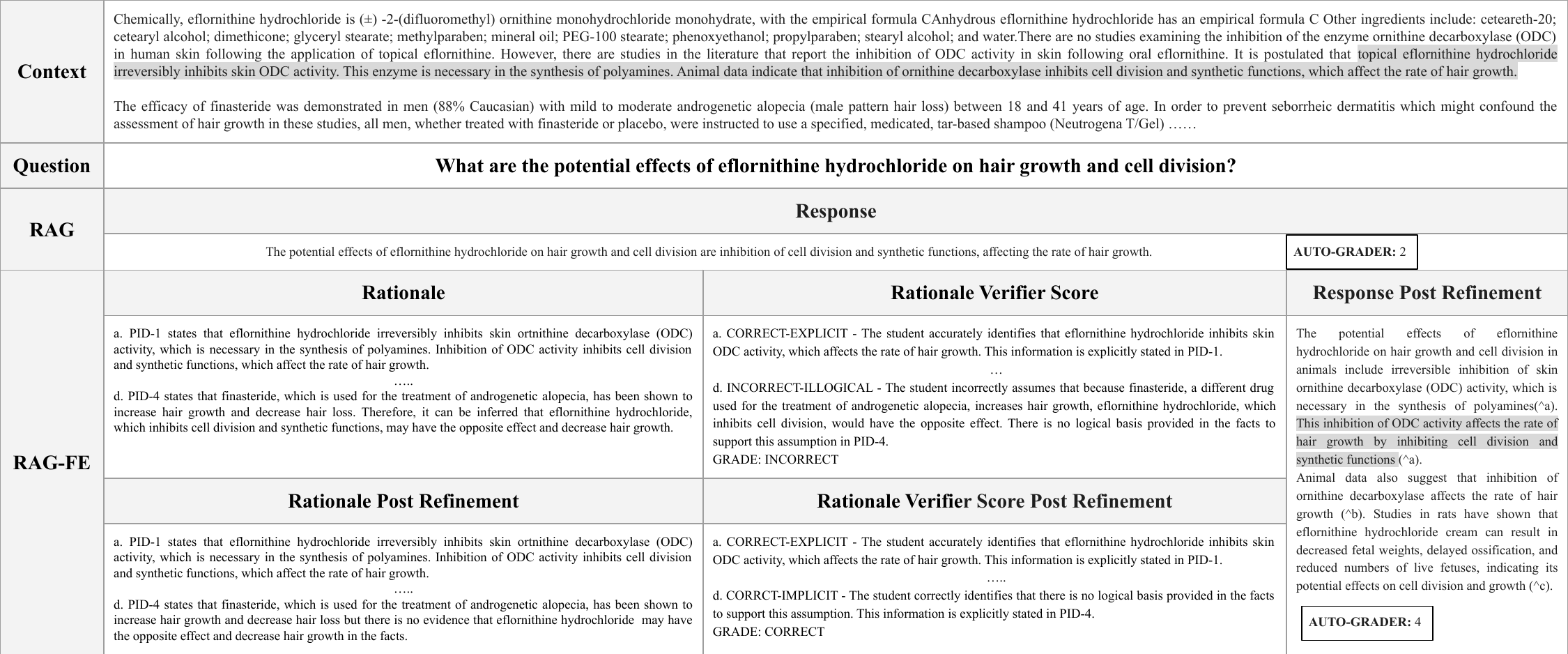}
    \caption{A qualitative analysis comparing RAG vs RAG+FE illustrates the superior quality of responses from RAG+FE}
    \label{fig: qualitative}
\end{figure*}

\subsection{Configuration} 
Hybrid Retrieval: The retriever was constant across all experiments, utilizing \textit{all-base-mpnet-v2}, a fine-tuned version of the model introduced by \cite{song2020mpnet}. This approach retrieves the top 5 paragraph chunks associated with the query.

LLM Generation: The following generation parameters were used for all methods and models. The temperature was set to 1, top\_p as 0.95, top\_k as 50 and do\_sample was False.

Finetuning: We fine-tuned LLama-2-7B on samples that were generated following the prompts by our proposed multi-stage framework.
The dataset consists of 32k long-form question-answering samples and is generated from the HotpotQA \cite{DBLP:journals/corr/abs-1809-09600} dataset. We utilized an effective batch size of 64, a learning rate of 3e-5, set the maximum sequence length to 4096 and trained the model for 2 epochs. The fine-tuning was powered by Deepspeed, employing 4 A100 GPUs. 

\subsection{Baseline}
We conducted a comparative analysis of our approach with two different methods, including Non-retrieval models (denoted as None), which lack contextual information and rely solely on pre-trained knowledge for query responses, and Retrieval-Augmented Generation (RAG) models that utilize contextual information in conjunction with the provided question.
Among the benchmark models considered were auto-regressive language models OpenAI GPT 3.5 Turbo and Meta Llama-2-7B, developed by Meta AI, employing a transformer architecture. 



\subsection{Metrics}
\subsubsection{Evaluation}
\label{evaluation}
Current evaluation methods such as n-gram \cite{lin-2004-rouge} and semantic-similarity based approaches \cite{zhang2020bertscore}, have limitations while assessing generative question-answering models. These methods fail to evaluate complex multi-hop reasoning QA, which is increasingly common as we move towards generative models from extractive ones. Recent works \cite{liu2023geval} have employed GPT-4 as the primary evaluator owing to their comparable performance to human evaluation. Since existing methods do not consider the rationale into account, we introduce Auto-Grader, a new evaluation technique that uses GPT-4 to systematically evaluate the relevance of rationale explanations and assign scores on a scale from 1 to 5. It also compares the generated answer to the ground truth answer, while providing detailed justifications for each assessment. This evaluation method is particularly useful for assessing generative models when their answers are not explicitly derived from the provided context. The Auto-Grader also evaluates the answer as either "CORRECT" or "INCORRECT", which can be used for calculating accuracy.
We provide the scale of grading below. Here the student refers to the generated LLM response (and is being graded by the teacher GPT-4):
\begin{enumerate}
    \item The student’s answer does not match both ground truth and context
    \item The student’s answer does not fully match the ground truth or the context, or it partially matches, and there may be irrelevant information
    \item The student’s answer matches the ground truth, but partially matches the context, and there may be irrelevant information
    \item The student’s answer matches with the ground truth and aligns well with the context, indicating a correct response within the given context
    \item  The student’s answer matches the ground truth, and it has taken additional relevant factual information about the key entities present in the query from the context, showcasing a deep understanding of the topic
\end{enumerate}
 
\subsubsection{Monitoring}
Monitoring real-world LLM applications is crucial to ensure that the model provides accurate answers. In real-time monitoring, the Rationale Verifier described in Section \ref{verifier} can be utilized to check whether the generated rationale adheres to the retrieved context. Additionally, for answer generation, the Faithfulness metric from RAGAS \cite{es2023ragas} can be used for monitoring purposes, as it only requires the query, context, and generated answer to provide a score. The generated answer is regarded as faithful if all the claims that are made in the answer can be inferred from the given context. This approach enables ongoing monitoring and ensures the continued accuracy of the system.

\begin{figure*}[t]
\includegraphics[width=0.85\textwidth]{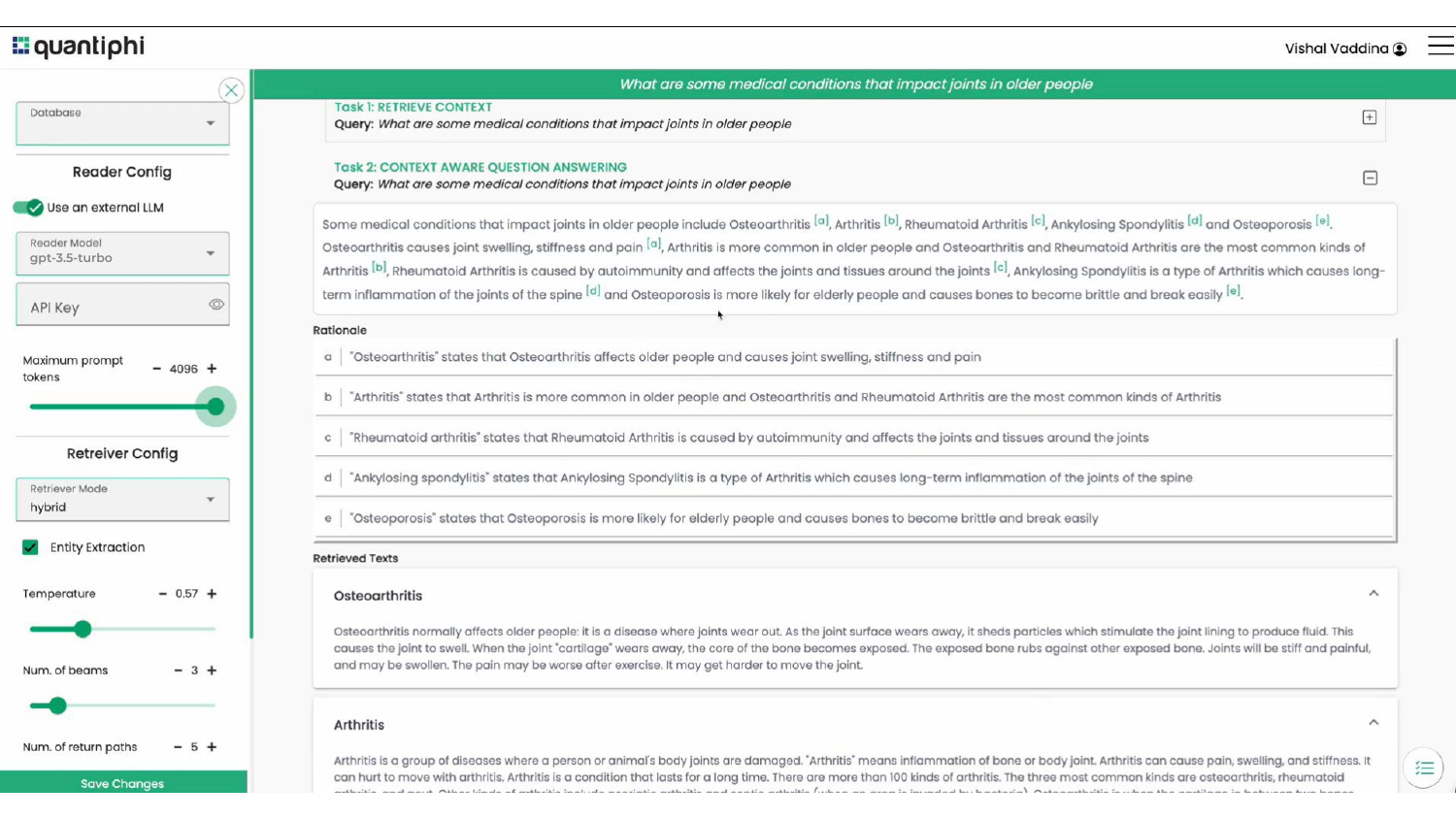}
  \caption{Illustration showcasing the interface, functionality, and overall presentation of the implemented system}
  \label{fig:arlabs}
\end{figure*}
\subsection{Performance Comparison}
Table \ref{tab:main-table} presents a comparison of the proposed framework (FE) with RAG and No Retrieval (None) on the PubMedQA and AEQA datasets, evaluated for Faithfulness, AutoGrading, and Accuracy metrics. The results show that RAG+FE outperforms RAG in terms of both faithfulness and accuracy across both datasets and models.

\textbf{Faithfulness:} Specifically, the highest faithful score achieved by RAG+FE is 86.50\% on the PubMedQA dataset, compared to 72.40\% for RAG. Similarly, on the AEQA dataset, RAG+FE achieves a faithful score of 83.24\%, which is significantly higher than RAG's score of 58.20\%.

\textbf{Accuracy}: RAG+FE achieves higher accuracy than RAG across both datasets and models. For instance, RAG+FE achieves an accuracy of 88.49\% on the PubMedQA dataset, compared to RAG's accuracy of 72.67\%, resulting in a relative improvement of 15.82\%. Similarly, on the AEQA dataset, RAG+FE achieves an accuracy of 96.85\%, which is significantly higher than RAG's accuracy of 75.19\%, with a relative improvement of 21.66\%. It is also observed that the accuracy computed by Auto-Grader is closer to the Faithfulness score, although the latter does not require the ground-truth answer for comparison.

The results also highlight the importance of the framework on smaller open-access models, where RAG + FE improves upon RAG. The scores of LLama-2-7B improve after fine-tuning and RAG + FE is comparable with the scores of vanilla RAG on GPT 3.5-turbo.

\textbf{Grade Score:} The main benefit of Auto-Grader is realized in its ability to grade based on the relevance and grounding of the response with the context. RAG+FE is better than RAG at effectively utilizing context in generating high-quality explanations. As mentioned in Section \ref{evaluation}, a score closer to 4-5 is ideal, while a score closer to 1 indicates an incorrect answer and a lack of effective utilization of context. RAG+FE achieves higher grades than RAG across both datasets and models, indicating its effectiveness in generating explanations that utilize context effectively.

\subsection{Qualitative Results} 

We provide a qualitative example of the different stages of our framework and a comparison with RAG in Figure \ref{fig: qualitative}.
We also provide a screenshot of our deployed system in Figure \ref{fig:arlabs}.


\section{Ablation Study}
\begin{table}[!ht]
\caption{Ablation Study on the effect of different components}
    \centering
    \begin{tabular}{p{4.2cm} c c}
    \hline
        Method & Accuracy (\%) & Faithful (\%) \\ \hline
        RAG (on Paragraphs) & 65.21 & 75.11 \\ 
        RAG + FE (Paragraphs w/ direct Rationale \& Answer generation) & 88.43 & 72.73 \\ 
        RAG + FE (Triplets instead of Paragraphs to reduce tokens) & 91.27 & 76.96 \\
        RAG + FE (Paragraphs w/o Rationale Verifier \& Refiner)  & 93.78 & 80.26 \\  
         RAG + FE (Paragraphs) & \textbf{96.85} & \textbf{83.24} \\ \hline
    \end{tabular}
    \label{table:ablation}
\end{table}
Table \ref{table:ablation} presents the results of an ablation study where different components of the RAG + FE approach were evaluated to assess their impact on accuracy and faithfulness. 

\textbf{FE outperforms vanilla RAG:} The results demonstrate that all variants of the RAG + FE approach outperform the baseline RAG method in terms of accuracy and faithfulness. 

\textbf{Impact of Rationale Verifier and Refiner:} Comparing paragraph retrieval experiments, the approach that uses the Rationale Verifier and Rationale Refiner achieves the highest accuracy of 96.85\% and the highest faithful score of 83.24\%. This suggests that verification and refinement lead to high-quality rationale and improve overall performance.

\textbf{Step-by-step vs direct generation:} Although the direct generation of both rationale and answer together is better than RAG alone, we observed better performance when done in a step-wise fashion, as there is a relative improvement of {8.42\%, 10.51\%} in terms of accuracy and faithfulness scores in terms of the latter.

\textbf{Paragraphs vs Triplets}: We experimented with triplets instead of paragraphs to reduce the context length of the input, reducing the average token length from 1846 to 285. While the number of tokens is reduced by 6.5x and can benefit downstream applications with limited computational resources, it has lower faithfulness and accuracy scores compared to the variant that uses paragraphs. Therefore, the choice of input representation should be made based on the specific requirements of the downstream application and the trade-offs between accuracy, faithfulness, and input length.

\section{Industry Impact}
Our work contributes to the ongoing efforts to enhance the reliability and credibility of LLMs in real-world industrial applications and extends beyond the previously demonstrated application in Pharmacovigilance. Domains where factual evidence for a response is essential include Legal, Finance (BFSI) and Education where the accuracy and transparency of a response is of utmost importance. Moreover, this framework alleviates the burden and time required by the user to double-check if a response is factually consistent. As this technology matures, its potential to revolutionize industries is evident, promising a safer and more efficient landscape for all stakeholders involved.

\section{Conclusion}
 We presented an approach to address the problem of hallucinations in LLMs by getting more accurate and plausible responses whilst providing a transparent explanation for the LLM's decisions. 
Furthermore, we have also demonstrated the applicability of our solution in the field of Pharmacovigilance, where ensuring the factualness of LLM responses is crucial. Through our quantitative and qualitative results, we have showcased how our approach can enhance the quality of responses for drug-related queries, such as identifying adverse drug reactions and providing precise and well-sourced explanations for the model's decision-making process. Our approach can be adopted by researchers and practitioners who are interested in combating factual hallucinations in LLMs. 

\bibliographystyle{ACM-Reference-Format}
\bibliography{sample-base}

\end{document}